\documentclass[12pt]{article}

\usepackage{arxiv}

\usepackage[utf8]{inputenc} 
\usepackage[T1]{fontenc}    
\usepackage{hyperref}       
\usepackage{url}            
\usepackage{booktabs}       
\usepackage{amsfonts}       
\usepackage{nicefrac}       
\usepackage{microtype}      
\usepackage{lipsum}
\usepackage{graphicx}
\usepackage{amsmath}
\usepackage{enumitem}
\usepackage{subcaption}
\usepackage{algorithm}
\usepackage{algpseudocode}
\usepackage{listings}
\usepackage{xcolor}
\usepackage[most]{tcolorbox}
\usepackage{listings}
\usepackage{tablefootnote}
\usepackage{hyperref}
\usepackage{multirow}
\usepackage{fontawesome5}
\definecolor{darkblue}{rgb}{0.0, 0.0, 0.5}
\hypersetup{
    colorlinks=true,
    urlcolor=darkblue,
}
\definecolor{shotOneColor}{RGB}{0, 51, 102}   
\definecolor{shotTwoColor}{RGB}{0, 102, 102}  
\definecolor{targetColor}{RGB}{153, 0, 76}    
\newcommand{\ourmethod}{\emph{Struct-SQL}}
\newcommand{\kdcot}{\emph{ReasonSQL}}
\newcommand{\qpcot}{\emph{QP-CoT}}
\newcommand{\texttosql}{\emph{Text-to-SQL}}
\newcommand{\fngold}{\emph{FN-Gold}}
\title{Knowledge Distillation with Structured Chain-of-Thought for Text-to-SQL}

\author{
Khushboo Thaker\textsuperscript{1},
Yony Bresler\textsuperscript{1}\\
\textsuperscript{1}Crater Labs, Toronto, Canada\\
\href{https://github.com/craterlabs/struct-sql-distillation}{\faGithub\ struct-sql-distillation} \quad
\href{https://huggingface.co/collections/craterlabs/struct-sql}{\includegraphics[height=0.9em]{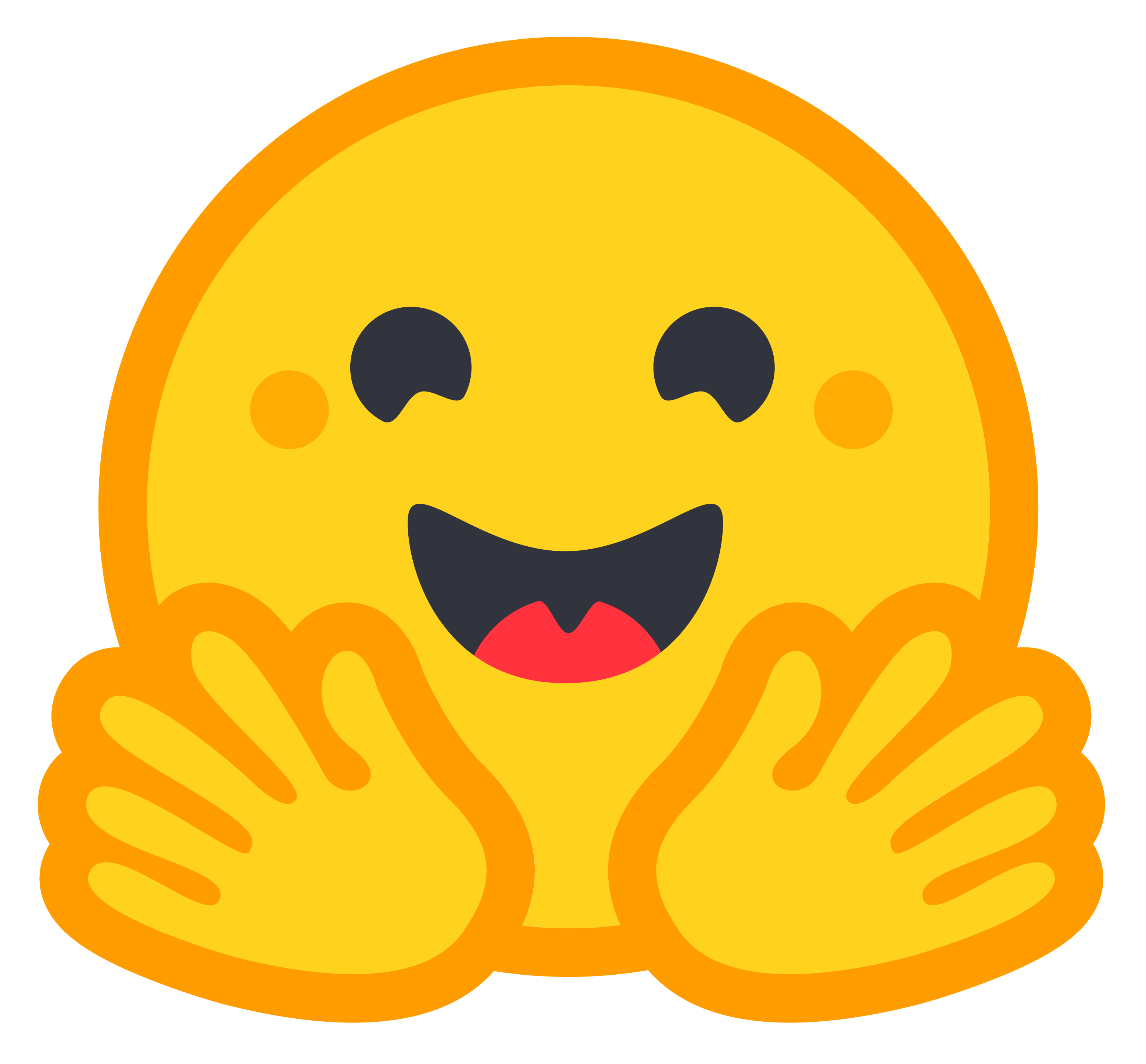} Struct-SQL}\\
\texttt{\{khushboo, yony\}@craterlabs.io}
}

\begin{document}
\maketitle

\begin{abstract}Deploying accurate \texttosql\ systems at the enterprise level faces a difficult trilemma involving cost, security and performance. Current solutions force enterprises to choose between expensive, proprietary Large Language Models (LLMs) and low-performing Small Language Models (SLMs). Efforts to improve SLMs often rely on distilling reasoning from large LLMs using unstructured Chain-of-Thought (CoT) traces, a process that remains inherently ambiguous. Instead, we hypothesize that a formal, structured reasoning representation provides a clearer, more reliable teaching signal, as the \texttosql\ task requires explicit and precise logical steps. To evaluate this hypothesis, we propose \ourmethod, a novel Knowledge Distillation (\textit{KD}) framework that trains an SLM to emulate a powerful large LLM. To implement this approach, we adopt the query execution plan as a formal blueprint to derive structured reasoning. Our SLM, distilled with a structured CoT, achieves an absolute improvement of 8.1\% over an unstructured CoT distillation baseline. A detailed error analysis reveals that a key factor in this gain is a marked reduction in syntactic errors. This demonstrates that teaching a model to reason using a structured logical blueprint is beneficial for reliable SQL generation in SLMs.
\end{abstract}
\section{Introduction}
\label{sec:introduction}

\texttosql\ (NL2SQL or text2sql) has the potential to democratize data access~\cite{fei2014constructing}. The field has seen substantial performance advancements driven by the advent of Large Language Models (LLMs)~\cite{shi2025asurvey}. These models enhance the capabilities of natural language interfaces for databases by automatically translating natural-language user questions into SQL queries~\cite{hong2024next}. Nevertheless, widespread adoption in enterprises remains challenging due to a difficult trade-off among three interdependent factors: cost, security, and performance. This challenge can be understood as an \textit{Adoption Trilemma}.
\begin{itemize}[leftmargin=*]
    \item Cost: High-performing models typically require significant computational resources, leading to high operational costs, whether using proprietary APIs or privately hosted LLMs~\cite{shi2025asurvey}.
   \item Security: Relying on external APIs raises serious security concerns, as transmitting potentially sensitive database schemas and sample records to third-party providers is often unacceptable in enterprise settings~\cite{hoang-etal-2025-distill}.
\item Performance: Selecting open-source models for local, private deployment to address cost and security issues often leads to the use of Small Language Models (SLMs), which typically lack adequate zero-shot accuracy for complex real-world queries~\cite{li2024can}.
\end{itemize}

To identify the limitations of current approaches, we examine the impact of recent advances in reasoning-driven prompting on \texttosql\ performance. A notable portion of the recent performance gains in \texttosql\ using LLMs can be attributed to in-context learning (ICL)~\cite{dong-etal-2024-survey}. ICL-based methods, particularly those that use decomposition and multi-step reasoning, have demonstrated substantial gains in performance on the \texttosql\ task \cite{hong2024next}. A prominent ICL technique for this is Chain-of-Thought (CoT), which encourages models to think step-by-step~\cite{wei2022chain, liu2025uncovering}. Such decomposition strategies are beneficial for generating complex SQL queries. For instance, \textit{DAIL-SQL}~\cite{gao2024dail} enhances CoT through context-aware examples, while \textit{DIN-SQL}~\cite{pourreza2023dinsql} and \textit{Divide-and-Conquer CoT} (\textit{DC-CoT})~\cite{pourreza2024chase} improve accuracy by breaking questions into intermediate sub-queries. Building on this logical foundation, the Query Plan CoT (\qpcot) guides the model through the steps that mirror a database execution plan, simulating the logical process by which databases execute queries. By following this structured path, \qpcot\ ensures that the SQL generation path aligns with the inherent logic of query execution~\cite{pourreza2024chase}. Although these reasoning techniques achieve considerable success, their effectiveness is observed almost exclusively in the large LLMs. This reliance on large LLMs constitutes a significant limitation: these methods depend on the very models that exacerbate the cost and security challenges, thereby limiting their applicability to resource-constrained enterprise settings and contributing to the `Adoption Trilemma'. 

The performance limitations of SLMs, suitable for private deployment, are particularly pronounced in the \texttosql\ domain. Standard benchmarks highlight a significant gap between large models and smaller open-source alternatives. For example, on the BIRD mini-dev benchmark, LLMs such as \textit{GPT-4o} and \textit{Claude 3.7 Sonnet} ($\geq$150B parameters) achieve execution accuracies between 30\% and 45\% using standard prompting, whereas widely used SLMs such as \textit{Mistral}, \textit{Mixtral} and \textit{Qwen2.5 Coder} ($\sim$7B parameters) achieve only 4\% to 12\%\footnote{Performance on BIRD mini-dev, see \url{https://github.com/bird-bench/mini_dev}}. This paper reveals that this severe performance degradation in SLMs persists even when advanced reasoning prompts are employed, despite their effectiveness in larger LLMs. This finding is consistent with recent literature, suggesting that large LLMs can adhere to the schema as long as the schema fits within their context window, a capability not observed in SLMs~\cite{maamari2024death}. In our experiments, the SLM (\textit{Qwen3-4B-Instruct-2507}) prompted with \qpcot\ fails primarily due to inadequate schema adherence, with a pronounced tendency toward schema hallucination, often generating non-existent tables or columns (see Figure~\ref{fig:pie_student}). Thus, while large LLMs benefit from structured reasoning strategies, SLMs do not internalize these logical decompositions. This breakdown in structured reasoning motivates an investigation into whether the \textit{structure of reasoning} itself can be effectively transferred from large LLMs to SLMs.

To bridge this performance gap, Knowledge Distillation (\textit{KD})~\cite{hinton2014distilling}, which transfers reasoning ability from a capable teacher model, is a key strategy~\cite{shi2025asurvey}. \emph{KD} is a model compression technique in which a smaller "student" model is trained to mimic the behavior of a larger, pretrained "teacher" model. Beyond compression, \textit{KD} enables the transfer of complex reasoning~\cite{li2024explanations} and logical skills~\cite{hsieh2023cotdistilling, yao2023specializing}. Through \textit{KD}, our aim is to build SLMs for \texttosql\ that deliver accuracy comparable to larger LLMs while meeting enterprise cost and security requirements through private deployment. This work argues that \textit{KD} is a promising approach to addressing this trilemma, but its effectiveness depends on the type of reasoning being distilled.

\begin{figure}[ht]
\centering
\centerline{\includegraphics[width=0.8\textwidth]{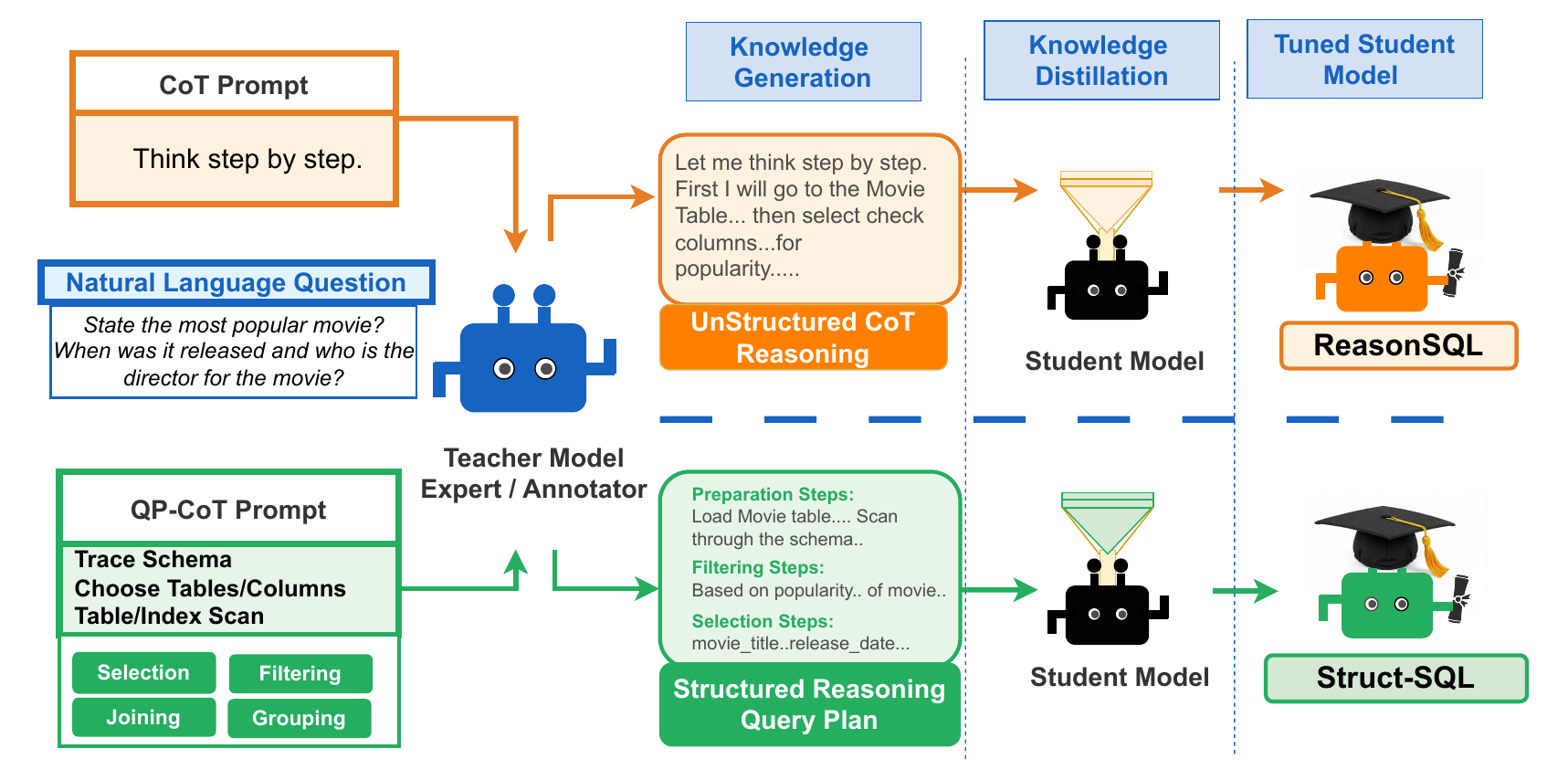}}
\caption{Unstructured vs. Structured Reasoning Distillation. The figure contrasts the two methods of reasoning distillation: (Top) Unstructured Distillation (\kdcot), which relies on a free-form CoT prompt, with (Bottom) the proposed Structured Distillation (\ourmethod), which uses a \qpcot\ prompt to generate a structured logical blueprint. The Teacher Model's output serves as the supervisory signal for tuning the Student Models.}
\label{fig:architect}

\end{figure}

Given that the core bottleneck for SLMs is replicating the teacher's complex reasoning, the central question for \textit{KD} is identifying the optimal form of the reasoning signal to distill. Building on the success of CoT prompting~\cite{wei2022chain}, one approach is to distill the teacher's natural language reasoning traces~\cite{li2024explanations}. This approach is exemplified by \kdcot, where the student model is trained on the 
teacher's intermediate CoT steps in addition to the final SQL query. This unstructured approach provides a better knowledge signal than finetuning on SQL queries alone and has been shown to improve model accuracy~\cite{rossiello2025rationalization}. However, we argue that the structure of the reasoning signal itself is critical for effective distillation. We hypothesize that distilling knowledge using a more formal, structured representation of the reasoning process that directly reflects the logical steps of query execution could provide a clearer, less ambiguous, and more beneficial supervisory signal for distilling SLMs than unstructured CoT explanations.

To evaluate this hypothesis, we introduce \ourmethod, a framework for distilling structured reasoning. Figure~\ref{fig:architect} shows the \textit{KD} workflow, contrasting the proposed structured reasoning approach against the standard unstructured CoT method. Within this framework, a state-of-the-art (SOTA) Teacher Model is used to generate a \qpcot\ trace, which formally decomposes the query into a logical execution plan \cite{pourreza2024chase}. This structured plan, together with the generated SQL query, constitutes the supervisory signal. A student model is then trained to replicate the entire structured output sequence (query plan, SQL). The formal query plan serves as a clear, hierarchical blueprint that guides the student model in learning the precise, logical steps of query construction, from schema linking and join-path selection to aggregation and filtering. During inference, this structured reasoning is retained: the student model is given the \qpcot\ prompt to autonomously generate the query plan before synthesizing the final SQL query.

To validate our hypothesis, we perform an extensive comparative analysis on the mini-dev BIRD benchmark~\cite{li2024can}, evaluating our \ourmethod\ framework against key baselines. Our main contributions are as follows.
\begin{itemize}
  \item We are the first to systematically evaluate the impact of \textit{KD} using a structured reasoning signal for \texttosql.
  \item We provide a comprehensive error analysis, showing that the improvement in \ourmethod\ results from a marked reduction in the syntactic errors (e.g., schema hallucination), demonstrating that the structured signal provides a clearer curriculum.
    \item We validate the generalization of our framework through an ablation study on two different SLMs.  
      \item We release the code (\url{https://github.com/craterlabs/struct-sql-distillation}) and, the model and the dataset (\url{https://huggingface.co/collections/craterlabs/struct-sql}) to facilitate reproducible research.
\end{itemize}
Following this introduction, Section~\ref{sec:methodology} details the methodology. Section~\ref{sec:experiments} presents the results, which are contextualized by the related work in Section~\ref{sec:related_work}, and the paper concludes in Section~\ref{sec:conclusion}. 

\section{Methodology}
\label{sec:methodology}

\subsection{Problem Formulation}
The standard \texttosql\ task involves mapping a natural language question $Q$ and a database schema $S$ to an executable SQL query $Y_\textrm{Gold}$. Formally, given an input pair $(Q, S)$, the goal is to learn a model $M$ such that $M(Q, S) = \hat{Y}$, where $\hat{Y}$ represents the predicted query intended to give the same result as the ground truth $Y_\textrm{Gold}$. Our work focuses on transferring the capabilities of a large, high-performance Teacher Model, $M_T$, to a smaller, more efficient Student Model, $M_S$, through \textit{KD}. We formulate the distillation task as learning the Teacher's intermediate reasoning steps, $R_T$, in addition to its final output, $Y_T$. Let $Z_T = R_T \oplus Y_T$ represent the complete output sequence. Training is conducted over a distillation dataset $\mathcal{D}_\textrm{DISTILL}$. We employ a standard sequence completion loss, minimizing the negative log-likelihood of the teacher-generated text $Z_T$ by updating the parameters $\theta$ of the student model:
\begin{equation}
\label{eq:lossfn}
  \mathcal{L}_{KD} = - \sum_{(Q,S,Z_T) \in \mathcal{D}_\textrm{DISTILL}} \log P_{M_S(\theta)}(Z_T | Q, S)
\end{equation}
We instantiate $R_T$ either as an unstructured CoT or a structured \qpcot\ to test the hypothesis that a structured reasoning trace provides a more effective supervisory signal.

\subsection{Structured CoT via Query Execution Plan}
\label{sec:qpcot}

\begin{figure}[ht] 
    \centering
    \includegraphics[width=.8\linewidth]{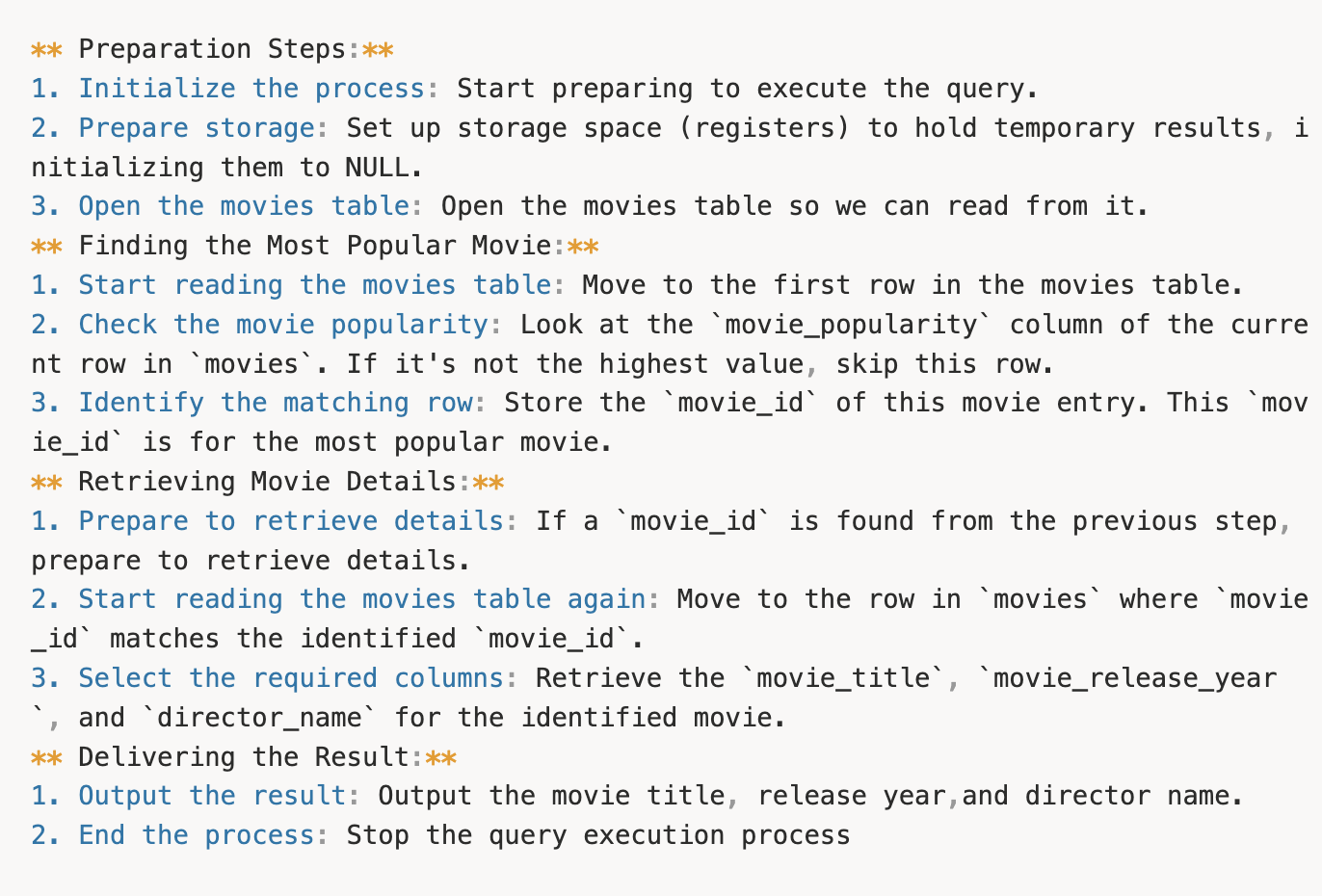} 
    \caption{ Sample structured query plan for the input "State the most popular movie? When was it released and who is the director for the movie?"}
    \label{fig:qp_plan}
\end{figure}

Our structured CoT strategy is inspired by a database engine's query execution plan. A query execution plan defines the precise sequence of steps that a database follows to access and manipulate data, often generated using an \textit{EXPLAIN} command. We adopt this ICL prompt strategy from a recent work on a Query Execution Plan-based prompting strategy,~\qpcot~\cite{pourreza2024chase}. Instead of a free-form explanation, the Teacher Model is prompted to generate a query plan along with the final SQL query. As shown in Figure~\ref{fig:qp_plan}, this plan decomposes the query into a sequential execution flow that explicitly performs selections, filters, joins, and groupings through step-by-step table scanning and data manipulation, providing a systematic signal for the student model. Although the query plan has been explored as a prompting technique~\cite{pourreza2024chase}, the key contribution here is to demonstrate the effectiveness of this structured reasoning as the primary teaching signal within a \textit{KD} framework. 

\subsection{Experimental Setup} 
\label{sec:assets}
The experiments use the SQLite-based~\textit{BIRD benchmark} for the data~\cite{li2024can}. The BIRD training dataset is used for model training, while the BIRD mini-dev dataset serves for evaluation. The Teacher Model ($M_T$), GPT-4o (OpenAI), is selected due to its leading performance on the BIRD mini-dev dataset and functions as an oracle to generate high-quality query plans. The \textit{Student Model} ($M_S$) is \textit{Qwen3-4B-Instruct-2507} (Alibaba Cloud). It was chosen for its strong performance-to-size ratio, which enables low-latency, private deployment and thus addresses the Adoption Trilemma~\cite{yang2025qwen3}. All models utilize a single-pass inference mechanism and operate without multi-agent collaboration, self-consistency checks or external correction loops. For a fair comparison with baselines, no external reasoning signals were provided at test time. During inference, the student model autonomously generates the Query Execution Plan based on the input prompt before generating the final SQL.

\subsubsection{Model Configurations}
    \noindent \textbf{Pretrained Models:} These configurations evaluate the model's intrinsic few-shot ICL capability using \qpcot\ without parameter updates.
\begin{itemize}
\item \textit{Teacher Model}: \textit{GPT-4o} ($M_T$) to establish the upper bound.
\item \textit{Student Model}: \textit{Qwen3-4B-Instruct-2507} to establish the lower bound. 
\end{itemize}

\noindent \textbf{Tuned Student Models:} These configurations evaluate the Student Model after tuning it on specific signals.
\begin{itemize}
\item \fngold\: The Student Model was finetuned using the Gold SQL query ($Y_\textrm{Gold}$) on the BIRD training dataset, with a basic system instruction that directly maps natural language to SQL~\cite{pourreza2023dinsql}.
\item \kdcot\ (unstructured \textit{KD}): The Student Model was finetuned on the complete Teacher sequence $Z_{T} = R_{CoT} \oplus Y_{T}$, where $R_{CoT}$ is the free-form CoT rationale~\cite{rossiello2025rationalization}. This configuration serves as the established baseline for evaluating the efficacy of standard unstructured reasoning distillation. 
\item \ourmethod\ (Structured \textit{KD}): The Student Model was finetuned on the entire Teacher sequence $Z_{T} = R_{\qpcot} \oplus Y_{T}$, where $R_{\qpcot}$ is the formal query plan CoT trace. This trace represents the structured logical blueprint designed for the core hypothesis. 
\end{itemize}
During inference, all models use the \qpcot\ prompt structure except for \kdcot, which employs the unstructured \textit{CoT} prompt.

\subsubsection{Distillation Dataset Construction}
\label{sec:data_generation}
The \textit{KD} datasets were constructed using an active generation and filtering methodology to maximize data quality and the diversity of query complexity. To initiate the process, the databases in the corpus are partitioned into a 75\% "in-domain" (ID) database pool and a 25\% "out-of-domain" (OOD) database pool, based on unique database identifiers. This partitioning ensures that the $\text{ID}$ pool serves as the source for all training data and the in-domain validation set, while the $\text{OOD}$ pool is used exclusively for the out-of-domain validation set, guaranteeing a robust measure of the trained model's generalization capabilities.

We then applied stratified, success-based sampling, where samples are grouped by SQL complexity categories and only admitted if the Teacher Model generates syntactically valid and execution-correct SQL. The ID corpus is iterated using predefined SQL structure and syntax categories to achieve target distributions for the training set (1,000 samples) and the in-domain validation set (150 samples). These categories, as shown in Table~\ref{tab:data_dist}, are differentiated by the complexity of the SQL structure, including single-table queries, queries with subqueries, queries with $\text{joins/set operations}$ and queries that combine both $\text{joins/set operations}$ and $\text{subqueries}$. For each candidate, inference was performed using \textit{GPT-4o}. A sample is admitted only if the generated SQL is both syntactically valid and yields the correct result upon execution. New samples are generated until 1,000 successful training samples and 150 ID validation samples are obtained. The OOD validation set (150 samples) is constructed using the same sampling and validation process, but applied exclusively to the segregated OOD database pool to ensure that no schema overlaps. For both \ourmethod\ and \kdcot, we used the same data generation pipeline to ensure a controlled, methodologically fair comparison. The datasets differ only in the format of the supervisory signal: unstructured CoT traces for \kdcot\ versus structured \qpcot\ for \ourmethod.

\begin{table}[ht]
  \centering
  \begin{tabular}{lrrr}
       \toprule
        \textbf{Complexity Category} & \textbf{Training Count} & \textbf{ID Val Count} & \textbf{OOD Val Count} \\
        \midrule
        Single Table Queries & 295 (29.50\%) & 37 (24.67\%) & 46 (30.67\%) \\
        Subquery (no join or set operations) & 229 (22.90\%) & 39 (26.00\%) & 31 (20.67\%) \\
        With JOINs / Set Ops (no Subquery) & 398 (39.80\%) & 57 (38.00\%) & 60 (40.00\%) \\
        JOINs / Set Ops and Subquery & 78 (7.80\%) & 17 (11.33\%) & 13 (8.67\%) \\
        \midrule
        \textbf{Total Successful Samples} & \textbf{1000} & \textbf{150} & \textbf{150} \\
        \bottomrule
    \end{tabular}
    \vspace{0.2cm}    
    \caption{Distribution of Query Complexity Across Distillation Datasets. The data is partitioned by unique database identifiers to evaluate generalization. ID Val = In-Domain Validation; OOD Val = Out-of-Domain Validation (databases unseen in training).}
    \label{tab:data_dist}   
\end{table}
\subsection{Implementation and Evaluation Details}
\label{sec:implementation}
\subsubsection{Post-Training Details}
\fngold, \kdcot, and \ourmethod\ were finetuned using Parameter-Efficient Fine-Tuning (PEFT) using Quantized Low-Rank Adaptation (QLoRA)~\cite{dettmers2023qlora} to improve training efficiency. \textit{PEFT} was chosen for its proven effectiveness in adapting SLMs to new domains with small training datasets. Furthermore, \textit{PEFT} has been shown to yield more stable models and mitigate the risk of catastrophic forgetting of foundational knowledge. \textit{QLoRA} allows recovery of near full 16-bit finetuning task performance even when the base model is loaded  at 4-bit precision~\cite{han2024parameterefficient}.
The \textit{KD} approach keeps the original model parameters $\theta_\textrm{BASE}$ fixed and only updates a small set of \textit{LoRA} adapter parameters $\theta_\textrm{ADAPTER}$. We followed the QLoRA methodology~\cite{dettmers2023qlora} and applied adapters to all linear layers of the Transformer architecture. Based on preliminary testing on a subset of the development data, we selected $r=64$ and an alpha scaling of $\alpha=128$ to ensure robust adaptation. Optimization was performed using AdamW with a learning rate of $10^{-4}$, a maximum input length of $15,000$ tokens, and a generation limit of $1,500$ tokens. We utilized a batch size of 6 for \kdcot\ and \ourmethod, compared to 15 for the \fngold\ baseline on an NVIDIA H200 GPU. All models minimized completion loss. To balance in-domain accuracy with out-of-domain generalization, we used an early-stopping strategy (patience=8) that monitored the aggregated validation loss across both the ID and OOD validation sets.
\subsubsection{Evaluation Metrics}
The primary metric is Execution Accuracy (EX), which measures the percentage of generated SQL queries that execute without error and return the same result set as the ground-truth SQL query on the BIRD mini-dev set. A detailed analysis of model failure modes reveals why and where certain models perform best. To facilitate analysis, we categorize failures into three distinct types as defined in Table~\ref{tab:failure_categories}. This classification establishes a severity hierarchy for \texttosql\ tasks: starting with the most severe Generation Failure (GEN), where the model produces no recognizable SQL output; followed by Syntactic Failure (SYN), which indicates an unexecutable query due to grammar or schema errors; and finally, Logical Failure (SEM), representing a query that is syntactically correct but semantically inaccurate. Recognizing this progression is beneficial for discerning the specific challenges and improvements associated with each model.

\begin{table}
    \centering
\begin{tabular}{@{}lll@{}}
    \toprule
    \textbf{Error Type} & \textbf{Subcategory} & \textbf{Description} \\
    \midrule
    Generation (GEN) & --- & No SQL output \\
    \midrule
    \multirow{5}{*}{Syntactic (SYN)} & No Such Column & Non-existent column reference \\
    & No Such Table & Non-existent table reference \\
    & Keyword Issue & Incorrect or misplaced SQL keyword \\
    & Syntax/Clause Order & Incorrect clause order, missing parentheses \\
    & Other &   Unclassified syntactic errors \\
    \midrule
    \multirow{5}{*}{Semantic (SEM)} & Column Mismatch & Incorrect number of columns \\
    & Row Mismatch & Incorrect number of row \\
    & Row \& Column Mismatch & Incorrect number of columns and rows \\
    & Value Mismatch & Incorrect data values \\
    & Empty Output & Empty result set \\
    \bottomrule
\end{tabular}
  \vspace{0.2cm}
    \caption{The proposed hierarchical error taxonomy. This table categorizes \texttosql\ errors in order of decreasing severity and is used for subsequent failure analysis.}
    \label{tab:failure_categories}
\end{table}

\section{Results}
\label{sec:experiments}

\subsection{Overall Performance}
The execution accuracy of \ourmethod, compared to all baselines on the BIRD mini-dev dataset, provides strong support for the central hypothesis. As summarized in Table~\ref{tab:main_results}, distilling structured reasoning results in significantly better performance compared to unstructured distillation, \kdcot\ and traditional finetuning \fngold. The native Student Model's performance (17.0\%) illustrates the Performance Trade-Off in the Adoption Trilemma, indicating that an SLM is insufficient for production deployment without intervention. \ourmethod\ achieved an $8.1$-point absolute improvement from $36.90$\% to $45.00$\% over the \kdcot\ baseline. This improvement enables the Student Model to cover 84\% of the Teacher Model's execution accuracy.
\begin{table}[ht]
  \centering
    \begin{tabular}{lc|rc|rrr}
        \toprule
     &  &  \multicolumn{2}{c|}{\textbf{Overall} } & \multicolumn{3}{c}{\textbf{EX by Difficulty} } \\
    \textbf{Model } & \textbf{Prompt} & \multicolumn{1}{c}{\textbf{EX}} & \textbf{Avg. Tokens} & \textbf{Simple} & \textbf{Moderate} & \textbf{Challenging}  \\
        \midrule
        Teacher (\textit{GPT-4o}) & \qpcot\ & 53.60\% & $298\pm96$& 68.24\% & 52.00\% & 36.27\%  \\
        Student Model & \qpcot\ & 17.00\%  & $1465\pm136$ & 34.45\% & 9.60\% & 9.80\% \\
        \midrule
        \multicolumn{6}{l}{\textbf{Tuned Student Models}}\\
        \fngold\ & \qpcot\ & 34.30\% & $198\pm257$ & 45.94\% & 32.80\% & 20.58\% \\
        \kdcot\ & CoT & 36.90\% &$99\pm99$ & 49.32\% & 33.20\% & \textbf{27.45}\% \\
        \kdcot\ (mismatched)\tablefootnote{Distilled using CoT, evaluated with QP-CoT.} & \qpcot\ & 29.20\% & $145\pm49$& 46.62\% & 24.80\% & 14.71\% \\

        \ourmethod\ (Ours) & \qpcot\ & \textbf{45.00\%} & $362\pm201$ & \textbf{65.54\%} & \textbf{40.4\%} & 25.49\%  \\
        \bottomrule
    \end{tabular}
    \vspace{0.2cm}          
    \caption{Overall and difficulty-wise execution accuracy (EX) on the BIRD mini-Dev set, along with average inference tokens (Avg. Tokens). \ourmethod\ achieves the highest EX among all tuned student models,  particularly for simpler and moderate queries. An $8.1$-point EX improvement over \kdcot\ requires $3.6$ times more tokens.}
    \label{tab:main_results}    
\end{table}

To isolate whether \ourmethod's performance gains stem from the structured \qpcot\ prompt format at inference time or from the structured supervisory signal during training, we conducted an additional ablation study. We evaluated the \kdcot\ model—trained on unstructured CoT traces—using the \qpcot\ prompt at inference, denoted as \kdcot\ (mismatched) in Table~\ref{tab:main_results}. This configuration achieved only 29.2\% execution accuracy, a substantial drop from the 36.9\% achieved when \kdcot\ uses its native CoT prompt. This 7.7-point degradation reveals a critical prompt-training mismatch: a model trained on unstructured reasoning cannot effectively leverage a structured prompt format, even when the prompt includes few-shot examples demonstrating how to generate the structured query plan. This suggests that in SLMs, \textit{ICL} alone is insufficient to bridge the gap; rather, the model must be explicitly trained on structured reasoning to internalize the required logical decomposition patterns. The strength of the training signal is further evidenced by \ourmethod, which achieves 45.0\% EX using the identical \qpcot\ prompt. This disparity validates that \ourmethod's success stems from internalizing the logical decomposition patterns during training, which the prompt alone cannot replicate.

\subsection{Analysis of Model Failures}
\begin{figure}[ht]
\centering
\begin{subfigure}[b]{0.48\textwidth}
    \centering
    \includegraphics[width=\textwidth]{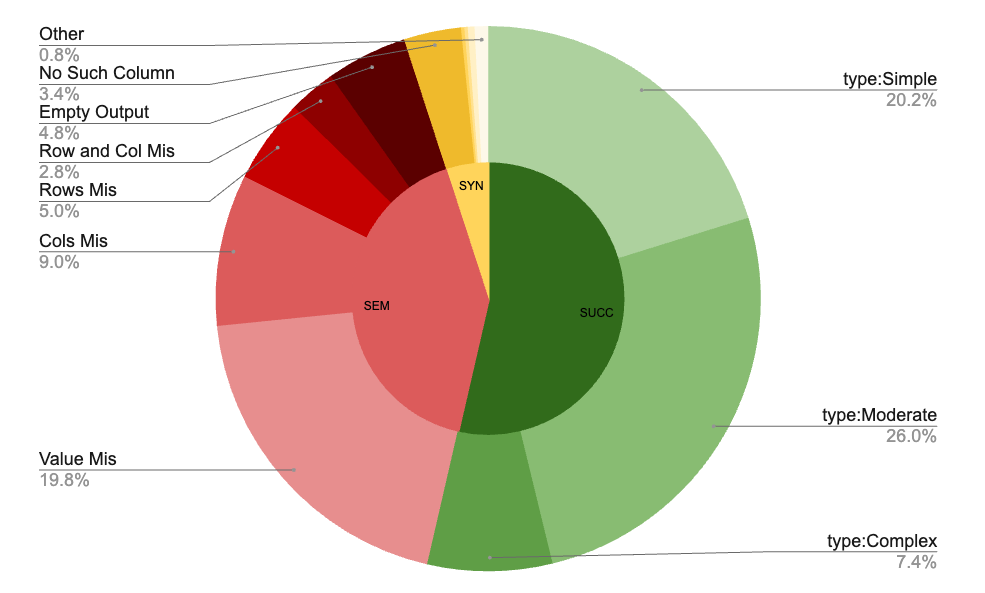}
    \caption{\textit{Teacher Model} using \qpcot}
    \label{fig:pie_teacher}
\end{subfigure}
\hfill
\begin{subfigure}[b]{0.48\textwidth}
    \centering
    \includegraphics[width=\textwidth]{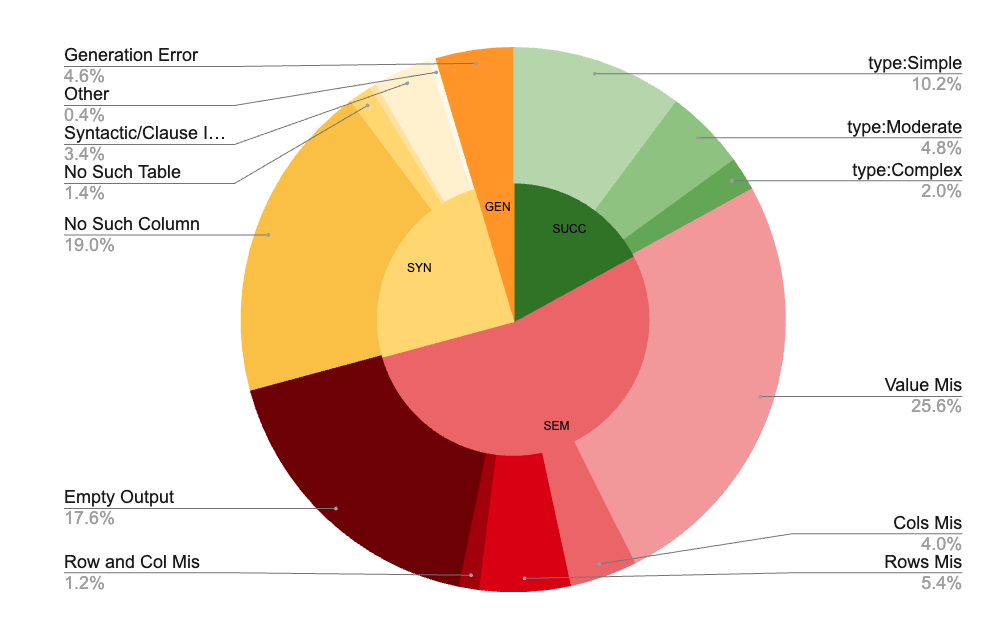}
    \caption{\textit{Student Model} using \qpcot}
    \label{fig:pie_student}
\end{subfigure}
\begin{subfigure}[b]{0.48\textwidth}
    \centering
    \includegraphics[width=\textwidth]{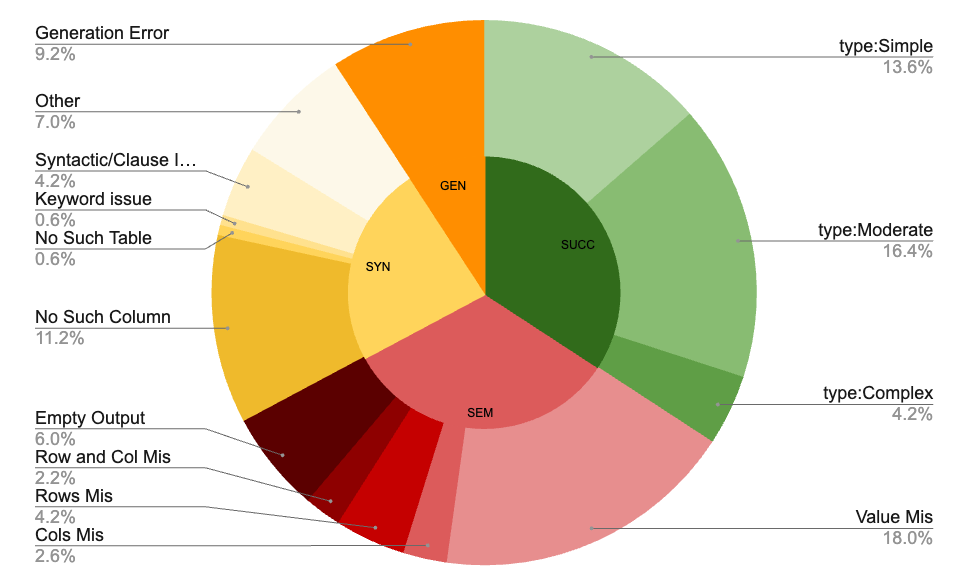}
    \caption{\fngold\ using \qpcot}
    \label{fig:pie_fn_gold}
\end{subfigure}
\hfill
\begin{subfigure}[b]{0.48\textwidth}
    \centering
    \includegraphics[width=\textwidth]{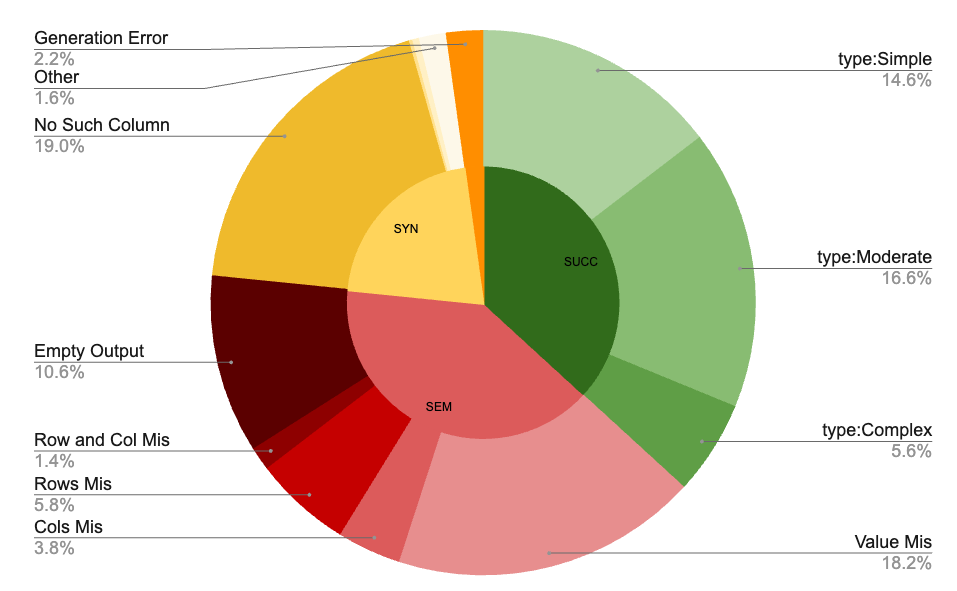}
    \caption{\kdcot\ using CoT}
    \label{fig:pie_kd_cot}
\end{subfigure}
\begin{subfigure}[b]{0.48\textwidth}
    \centering
    \includegraphics[width=\textwidth]{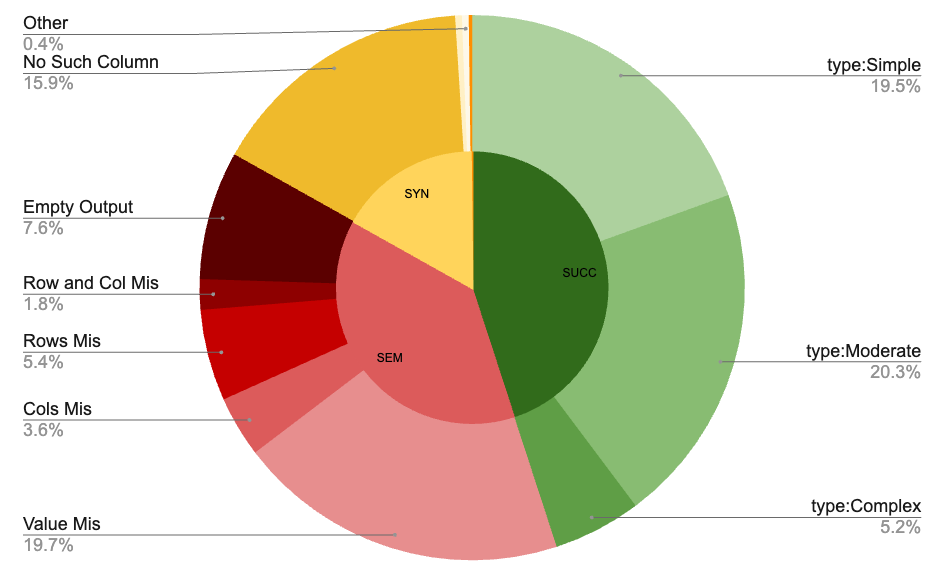}
    \caption{\ourmethod\ using \qpcot}
    \label{fig:pie_kd_qp_cot}
\end{subfigure}
\centering

\caption{Compared to the Teacher Model (a), both the Student Model (b) and the \fngold\ (c) exhibit substantially lower performance, primarily due to high syntactic errors. The unstructured distillation baseline \kdcot\ (d) improves upon both the Student Model and \fngold. \ourmethod\ (e) achieves the highest success rate among all tuned student models. }
\label{fig:pie_charts}

\end{figure}

Figure~\ref{fig:pie_charts} presents hierarchical pie charts. The inner ring displays top-level results that depict success versus categorized error types, while the outer rings further specify subcategories of failures, as defined in Table~\ref{tab:failure_categories}. For successful queries, the outer ring categorizes results by the difficulty level of the SQL queries in the BIRD dataset.
As visually summarized in the hierarchical pie charts, a key dichotomy emerges between the Teacher and other baselines, primarily distinguished by their dominant error types. The Teacher Model ($M_T$), shown in Figure~\ref{fig:pie_teacher}, achieved the highest success rate. Its failures were predominantly semantic errors ($\sim$30\%), indicating challenges with logical complexity. The Student Model ($M_S$), shown in Figure~\ref{fig:pie_student}, had a higher number of syntactic errors, including schema hallucination, underscoring a fundamental inability to adhere to the database schema before post-training. 

The standard finetuning baseline (\fngold) reflects the performance achieved by training solely on the gold SQL($Y_\textrm{Gold}$), without incorporating intermediate reasoning. As shown in Figure~\ref{fig:pie_fn_gold}, this approach focused its gains almost entirely on reducing semantic errors (from 53.8\% to 33.0\%, a $20.8$-point drop) of the Student Model $M_S$, but failed to address the syntactic bottleneck: overall syntactic errors remained nearly unchanged ($24.6\%$ to $23.6\%$). These results confirm that training on only the final query output is insufficient for learning strict SQL grammar. 

The unstructured distillation (\kdcot) baseline evaluates distillation through an unstructured reasoning trace ($R_{CoT}$). As shown in Figure~\ref{fig:pie_kd_cot}, this approach achieved an execution accuracy of $36.9\%$, higher than that of $M_S$ and \fngold, indicating that unstructured \textit{CoT} reasoning provides a better distillation signal than both the naive model and \fngold. \kdcot\ significantly reduced the rate of generation errors compared to the naive Student Model $M_S$, suggesting that unstructured distillation enhanced the stability of the model. Furthermore, it reduced semantic errors from 53.8\% to 39.8\% (a $14.0$-point reduction) and reduced syntactic errors from $24.6\%$ to $21.2\%$ (a $3.4$-point reduction), primarily by improving basic schema linking and eliminating `No Such Table' errors. These findings align with previous research demonstrating that CoT-based distillation improves accuracy~\cite{hsieh2023cotdistilling} and further clarify its specific advantages.

As established above, unstructured distillation \kdcot\  outperformed both the naive Student Model and standard finetuning \fngold. \ourmethod, as shown in Figure~\ref{fig:pie_kd_qp_cot}, outperformed \kdcot. By distilling the reasoning signal into a structured blueprint, \ourmethod\ increased execution accuracy from $36.9\%$ to $45.0\%$ (an $8.1$-point improvement) compared to \kdcot. This improvement appeared across the error severity hierarchy. \ourmethod\ nearly eliminated generation errors, reducing the rate from $2.2\%$ to $0.4\%$ (a $1.8$-point reduction). For syntactic errors, although \kdcot\ outperformed the naive model, it did not enforce strict schema alignment. In contrast, \ourmethod\ reduced the total syntactic error count from $21.2\%$ in \kdcot\ to $16.8\%$ (a $4.4$-point reduction). This reduction included fewer `No Such Column' hallucinations, $19.0\%$ to $15.8\%$ (a $3.2$-point reduction) and the elimination of `Keyword Issues', showing a more precise understanding of SQL syntax. For semantic errors, the structured model showed greater logical reliability, reducing `Empty Output' errors from $10.6\%$ in \kdcot\ to $7.6\%$ (a $3.0$-point reduction). Although this increased rigidity led to a minor trade-off, \kdcot\ retained a slight advantage in `Value Mismatches' ($18.2\%$ vs. $19.6\%$), suggesting that the structured plan provides a more robust signal.
%

\subsection{Fine-Grained Performance Analysis and Ablation Study}
To investigate the precise source of performance of \ourmethod, we move beyond aggregate metrics to examine the detailed differences between models.
In this section, we deconstruct the model's capabilities across three dimensions: its proficiency with complex SQL operations (Figure~\ref{fig:heatmap}), the fidelity of its alignment with the teacher model (Figure~\ref{fig:gains_losses}), and its ability to systematically rectify the student model's baseline errors (Figure~\ref{fig:sankey}).

\subsubsection{Performance by SQL Construct:} 
\begin{figure}[ht]

\centering
\begin{subfigure}[b]{0.45\textwidth}
    \centering
    \includegraphics[width=\textwidth]{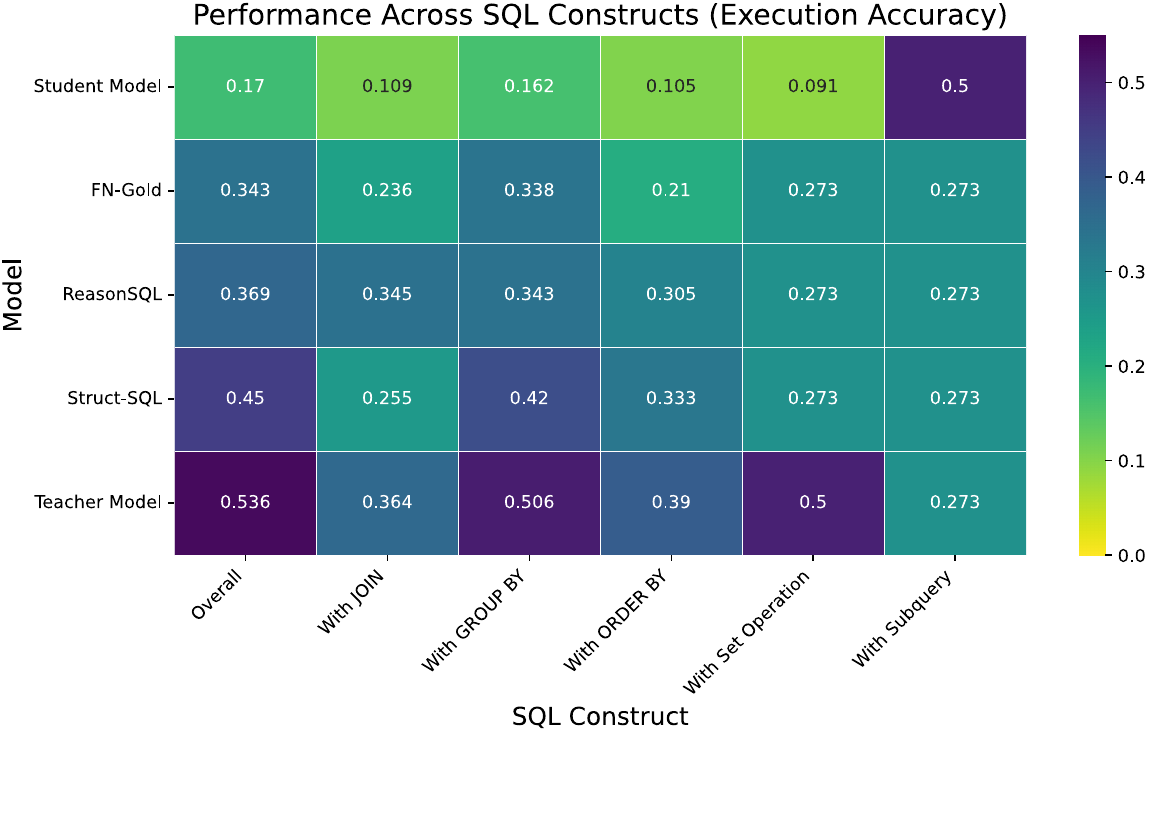}
    \caption{\text{EX by SQL Construct}}
    \label{fig:heatmap}
\end{subfigure}
\hfill
\begin{subfigure}[b]{0.45\textwidth}
    \centering
    \includegraphics[width=\textwidth]{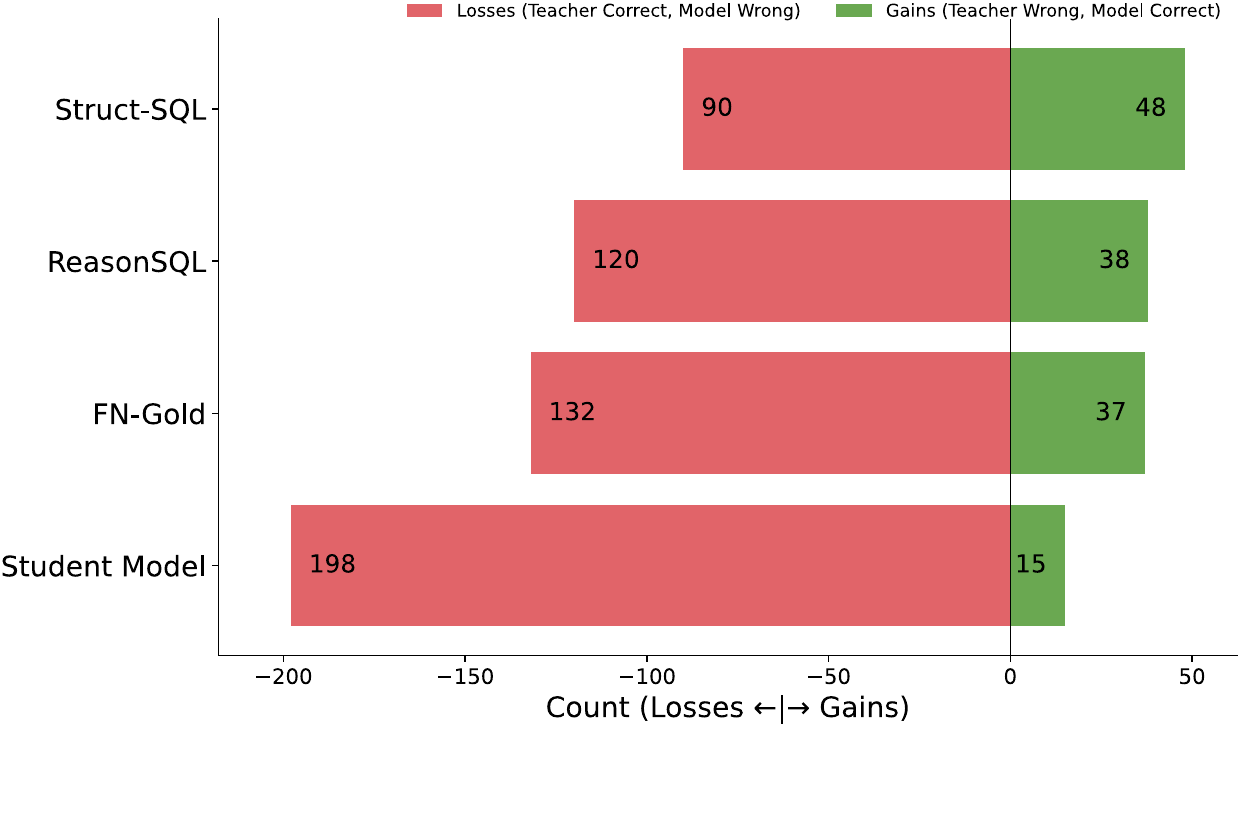}
    \caption{\text{Gains vs. Losses Analysis}}
    \label{fig:gains_losses}
\end{subfigure}
\begin{subfigure}[b]{0.7\textwidth}
    \centering
\includegraphics[width=\textwidth]{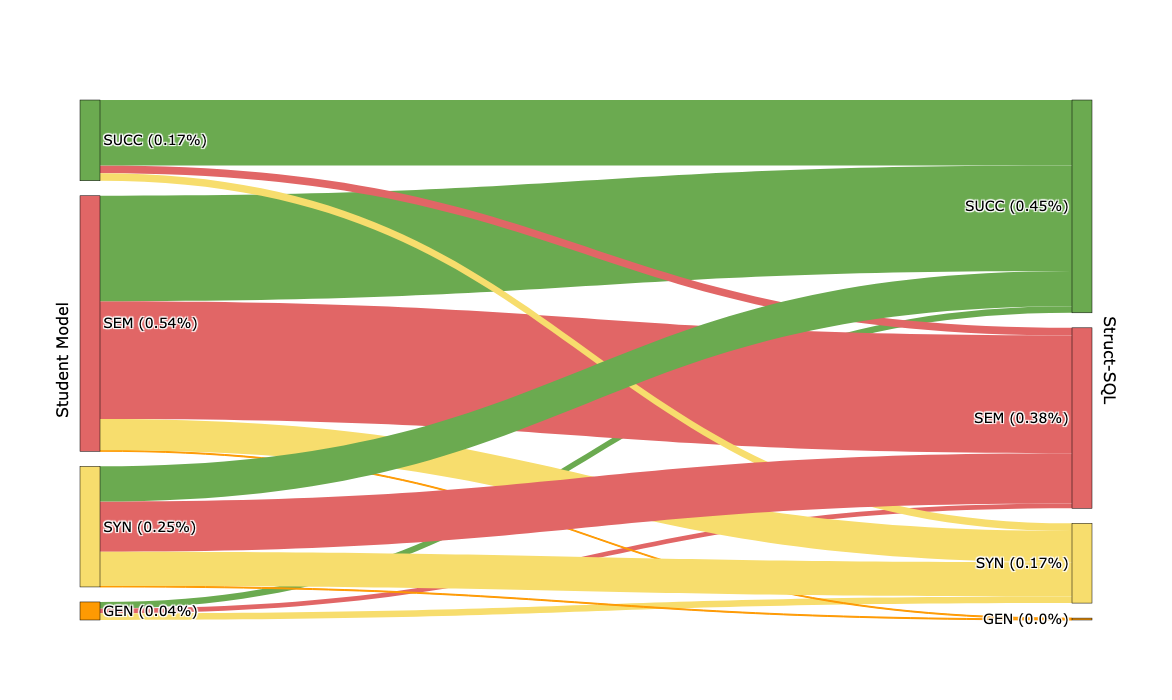}
\caption{\text{Performance State Transitions} }
    \label{fig:sankey}
\end{subfigure}
\begin{minipage}[b]{1\textwidth}
        \centering
        \text 

\end{minipage}

\caption{Detailed performance analysis. (a) Execution Accuracy across different SQL constructs, highlighting \ourmethod's proficiency in handling complex aggregations. (b) Gains vs. Losses analysis for baseline models relative to the Teacher model. The losses and gains are correlated with overall performance. 
(c) Performance State Transitions illustrates \ourmethod's effectiveness in converting severe errors (SYN, GEN) from the Student Model into  direct successes or less severe errors (SEM).}
\label{fig:performance}

\end{figure}

Assessing model performance across the spectrum of query formulations can be insightful; therefore, we analyzed execution accuracy for key SQL constructs, as shown in Figure~\ref{fig:heatmap}. The comparative analysis indicates that $\ourmethod$'s performance gains are most pronounced in tasks that require aggregation and explicit structural decomposition. This is evidenced by its strong results on specific SQL operations. For queries requiring a GROUP BY clause, the $\ourmethod$ model ($42.0\%$ EX) outperformed both \fngold\ ($33.8\%$ EX) and \kdcot\ ($34.3\%$ EX), demonstrating that explicitly distilling the query execution plan effectively trains the model on the necessary aggregation formulation. An advantage was also observed for queries requiring an ORDER BY, where $\ourmethod$ showed a $2.8$-point absolute improvement over \kdcot. However, the comparative analysis revealed a notable exception: for queries that require $\text{JOIN}$ operations, \kdcot\ achieved the highest execution accuracy. In contrast, \ourmethod's score ($25.5\%$ EX) was lower, suggesting that the unstructured $\text{CoT}$ trace provided a unique advantage for learning multi-entity linking. This area remains a core challenge for all models, as confirmed by the Teacher Model's performance ($36.4\%$ EX). Furthermore, all baselines, including \ourmethod, shared a weakness in handling queries that require Set Operations.

\begin{table}[ht]
\centering
\begin{tabular}{l|c|ccc}
\toprule
 & \textbf{ Overall} & \multicolumn{3}{c}{\textbf{ EX by Difficulty}} \\
\textbf{Model} & \textbf{ EX (\%)} & \textbf{ Simple} & \textbf{ Moderate} & \textbf{ Challenging} \\ 
\midrule
Student Model & 7.22 & 11.49 & 6.00 & 3.92 \\
\kdcot\ & 25.10 & 40.54 & 20.08 & 12.75 \\
\ourmethod\ (Ours) & \textbf{29.31} & \textbf{49.32} & \textbf{23.20} & \textbf{14.71} \\ 
\bottomrule
\end{tabular}%
\vspace{0.2cm}    
\caption{Ablation study on \textit{Mistral-7B-Instruct-v3.0} demonstrates that \ourmethod\ continues to outperform unstructured \kdcot\ across different base models.}
\label{tab:mistral_ablation}

\end{table}

\subsubsection{Evaluation of Knowledge Transfer: Gains vs. Losses:} A Gains vs. Losses analysis was conducted to assess the quality of the distilled knowledge (Figure~\ref{fig:gains_losses}). This approach compares queries where the Teacher was correct, but the Student Model failed (Losses, red segments) with cases where the Student Model was correct, but the Teacher Model was incorrect (Gains, green segments), representing true generalization. The results indicate that $\ourmethod$ demonstrates the highest net performance relative to the Teacher Model  by exhibiting the fewest losses and the most gains among all student models. This confirms that the structured reasoning signal leads to better fidelity in knowledge transfer.

\subsubsection{Performance State Transitions:} Figure~\ref{fig:sankey} illustrates the progression of queries from the baseline Student Model to the distilled 
\ourmethod, highlighting the systematic conversion of error states into successful or unsuccessful output. \ourmethod\ retained more than 81\% of the initial successes and enhanced stability by reducing generation errors of the Student Model from 4.0\% to 0.4\%. The structured framework also corrected 41.3\% of semantic errors (SEM $\to$ SUCC), confirming that the formal query plan addresses logical flaws. Additionally, over two-thirds of syntactic failures were upgraded: 29\% were corrected (SYN $\to$ SUCC) and 41\% shifted to semantic errors (SYN $\to$ SEM).

\subsection{Generalization on \textit{Mistral-7B}:} To validate the transferability of our framework to other architectures, we replicated our experiments on \textit{Mistral-7B-Instruct-v3.0}. This model was specifically selected for its low zero-shot accuracy of 7.2\% EX on BIRD mini-dev with \qpcot\ prompt. As shown in Table~\ref{tab:mistral_ablation}, \ourmethod\ shows generalizability, achieving a performance advantage (29.3\% EX) over the Student Model (7.2\% EX) and the \kdcot\ baseline (25.1\% EX), confirming that the structured query plan offers a robust supervision signal independent of the base model.

\subsection{Computational Efficiency:} \label{sec:computational_efficiency}
On a single H200 GPU, using \textit{Qwen3-4B-Instruct-2507} as the Student Model, \ourmethod\ converged using an early stopping strategy with a patience of 8 steps and a threshold of 0.001 in 2.24 epochs, requiring only 29.15 minutes. This training efficiency is comparable to that of the unstructured \kdcot, which required 25.24 minutes across 6.4 epochs with the same batch size, sample count and early stopping strategy. This 1,000-sample distillation strategy is also computationally tractable compared to traditional finetuning on larger datasets. For context, the \fngold\ baseline, trained on the entire 9,000+ sample BIRD dataset, required 110.57 minutes (4.33 epochs) to converge. This efficiency underscores the practical deployability of our structured distillation method in resource-constrained environments.

\subsection{Official Test Evaluation:} As shown in Table~\ref{tab:bird_official_test}, \ourmethod\ obtained 60.42\% EX on the non-public BIRD benchmark~\cite{li2024can}, securing the top position among $\leq$4B models (as of January 30, 2026) in single model inference. This result is achieved using strict single-model inference with greedy decoding and no self-consistency.

\begin{table}[h]
\centering
\caption{Official BIRD Bench Evaluation Results\tablefootnote{https://bird-bench.github.io/}. Struct-SQL (4B) ranks first globally among models with $\leq 4$B parameters (as of January 30, 2026).}
\label{tab:bird_official_test}
\begin{tabular}{lcccc}
\toprule
\textbf{Model} & \textbf{Simple (\%)} & \textbf{Moderate (\%)} & \textbf{Challenging (\%)} & \textbf{Overall EX (\%)} \\
\midrule
\ourmethod (4B) & 69.02 & 54.41 & 43.51 & 60.42 \\

\bottomrule
\end{tabular}
\end{table}

\section{Related Work}
\label{sec:related_work}

\subsection{\texttosql\ with In-Context Learning (ICL)} LLMs fundamentally redefined the field of \texttosql~\cite{deng2022recent}, achieving substantial performance on challenging cross-domain benchmarks such as  Spider~\cite{yu2018spider} and BIRD~\cite{li2024can}. ICL~\cite{dong-etal-2024-survey} has emerged as a dominant research paradigm, enabling LLMs to leverage their pre-training by processing instructions and examples for SQL generation~\cite{shi2025asurvey}. Initially, ICL approaches utilized foundational zero-shot and few-shot paradigms~\cite{dong2023zero, rajkumar2022evaluating, liu2023comprehensive, wang-etal-2025-mac}. However, generating complex SQL queries that involve nested logic and multitable joins was found to require explicit external guidance~\cite{pourreza2023dinsql}. As a result, the field pivoted toward decomposition, a multi-step reasoning tactic designed to improve accuracy and reliability~\cite{pourreza2023dinsql, zhang2023act, gao2024dail, Li2024pet, liu2023divide}. As a form of ICL-based decomposition, basic CoT helped LLMs break down the task, follow a logical workflow, and improve execution accuracy~\cite{liu2023divide}. This progression led to sophisticated variants such as \textit{DIN-SQL}~\cite{pourreza2023dinsql}, which decomposes the task into sequential stages (e.g., schema linking, query classification). Additional multi-stage architectural interventions have also been explored~\cite{zhang2023act, Li2024pet}. Crucially, these approaches often require multiple LLM calls to intervene, correct, or gather intermediate data, leading to high latency and cost. To mitigate this, some approaches focus on optimizing efficiency by creating single-pass Structural CoT approaches~\cite{pourreza2024chase, liu2023divide}. These methods instruct the LLM to follow a formalized logical blueprint to gain the necessary understanding of the schema and natural language before generating the SQL. Specialized methods, including \qpcot\ and \textit{QDecompose}, excel here, as they have been shown to produce better results than non-structured CoT~\cite{pourreza2024chase, gao2024dail}. For example, \qpcot\ provides few-shot query plan examples that require the model to first generate a detailed query plan before generating SQL code~\cite{pourreza2024chase}. Although single-pass ICL performs well with large LLMs, its gains are limited when applied to SLMs ~\cite{maamari2024death} or require expensive finetuning~\cite{pourreza2024dts, liu2025uncovering, gorti-etal-2025-msc, zhai2025excot}. This work addresses this identified gap by introducing \ourmethod, a novel \textit{KD} method that enables single-pass ICL for SLMs.

\subsection{\textit{KD} for Reasoning Transfer}
$\textit{KD}$ has evolved from a model compression technique to a method for complex knowledge transfer, called Skill Distillation, which is often used to transfer reasoning knowledge from one model to another or to facilitate self-improvement~\cite{hinton2014distilling, shi2025asurvey}. Due to the significant resource demands of LLM, recent research has focused on efficiently transferring their reasoning capabilities from a teacher LLM to a student SLM. This transfer is frequently grounded in \textit{ICL} principles, demonstrating that learning from few-shot demonstrations can be successfully distilled into the SLMs' parameters~\cite{snell2022learning}. The primary established approach to enable reasoning in SLMs is the Finetune-CoT paradigm~\cite{li2024explanations}, which uses unstructured CoT explanations from the teacher as a supervision signal. This distillation approach has been demonstrated  to be effective with multi-step mathematical reasoning~\cite{yao2023specializing}. However, relying on unstructured CoT often leads SLMs to learn spurious correlations rather than deep causal features, thereby reducing robustness on complex data~\cite{li2023mixed, shridhar2023distilling}. Consequently, recent research increasingly favors methods that enforce structural decomposition and explicit causality. These approaches include creating constrained distillation pipelines with the aim of removing ambiguous input context through advanced structural formats. For example, $\textit{SocraticCoT}$ provides an explicit structural format by guiding the model through reasoning using defined subquestion-solution pairs~\cite{shridhar2023distilling}. Alternatively, $\textit{Mixed Distillation}$ integrates the CoT with a formal verifiable Program-of-Thought reasoning to provide a less ambiguous supervision signal~\cite{li2023mixed}. In the \texttosql\ domain, structured \textit{KD} has seen limited uptake. Previous studies, while validating the effectiveness of distillation in this area, have generally relied on less structured signals, such as schema-based finetuning to improve schema linking~\cite{hong2024next}, the inclusion of enterprise-specific custom examples~\cite{hoang-etal-2025-distill}, and unstructured CoT reasoning transfer~\cite{rossiello2025rationalization}. The proposed $\ourmethod$ framework systematically evaluates whether structure-based reasoning transfer, specifically \qpcot, improves the distillation for \texttosql. 

\section{Limitations}
\label{sec:limitations}

\ourmethod\ significantly reduces syntactic errors through the use of structured reasoning; however, several constraints remain:

\begin{itemize}
    \item \textbf{Token Overhead and Efficiency:} \ourmethod\ significantly reduces syntactic errors through structured reasoning; however, this approach introduces inference overhead. The generation of an intermediate query plan requires approximately 3.6 times more tokens than \kdcot\ (Table~\ref{tab:main_results}). This increase impacts latency and operational costs; nevertheless, these requirements may still be lower than those of teacher LLMs. It would be worthwhile to investigate more concise query plan templates to explore this trade-off.
    

    \item \textbf{Performance Bound:} The student model's capabilities are fundamentally capped by the teacher's fidelity. Our distillation pipeline utilizes a strict filtering mechanism, training only on instances where the teacher model generates an execution-correct SQL query. Although this quality control is essential to prevent error propagation and provide a stable signal for small models~\cite{zhang2025can}, it inherently skews the training data toward problems that the teacher can solve, potentially limiting the student's exposure to edge cases that the teacher itself cannot solve~\cite{qian2025good}. As a result, even with a competitive 60.42\% EX on the BIRD test set, the performance remains bound by the teacher's own limitations in handling highly complex multi-level joins and set operations, as discussed in Figure~\ref{fig:heatmap}. To improve performance further, future work could explore using multiple teachers or human-annotated data to cover edge cases that a single source model may miss.
    
    \item \textbf{Benchmark Selection:} A fixed reasoning template is used to minimize syntactic hallucinations through a constrained logical path; however, we recognize that it may limit flexibility across highly diverse database dialects. We intentionally prioritize evaluation on the BIRD benchmark, as it represents a more challenging dataset than SPIDER. As demonstrated in recent literature (e.g., Snowflake Arctic-Text2SQL-R1), state-of-the-art models frequently exceed 88\% EX on the SPIDER-test, yet often struggle to surpass 70\% on BIRD-dev \cite{yao2025arctic}. This performance gap justifies our decision to focus exclusively on BIRD.
\end{itemize}

\section{Conclusion and Future Research}
\label{sec:conclusion}
This work presents \ourmethod, a \textit{KD} framework that transfers structured reasoning (\qpcot) from a Teacher LLM to a smaller student model for the \texttosql\ task. The central hypothesis is that a formal, structured reasoning signal provides a superior, less ambiguous teaching blueprint than unstructured reasoning. The experiments confirm the hypothesis, demonstrating that distilling structured reasoning is a better teaching method. Detailed error analysis indicates that these improvements are primarily due to a significant reduction in syntactic errors. An additional ablation study demonstrated generalization to a different base model. Given the demonstrated effectiveness of structured signals in the refinement of syntactic efficiency and logical reasoning, further research is warranted to validate the effect of this structured distillation approach on other complex reasoning tasks beyond \texttosql. In summary, the proposed framework facilitates the deployment of high-performing, cost-effective and private models.

\section{Acknowledgments}
We thank Khalid Eidoo, Abdul Hamid Dabboussi, Mar\'ia Rodr\'iguez-Li\~n\'an and Ting Fung Lam for their insightful technical discussions and valuable feedback throughout the project and during the review of this manuscript.

\bibliographystyle{unsrt}  
\bibliography{references} 
\newpage
\appendix

\section{Structured Reasoning Prompt Template (QP-CoT)}
\label{sec:appendix_prompt}

The \qpcot\ prompt, as detailed in Listing~\ref{lst:qp_cot_prompt}, comprises a system instruction followed by two few-shot examples that demonstrate the query execution plan methodology. The prompt concludes with the target task; for the sake of brevity in this manuscript, the database schema definitions have been abbreviated.
\begin{lstlisting}[
    basicstyle=\ttfamily\scriptsize,
    breaklines=true,
    frame=single,
    backgroundcolor=\color{gray!5}, % Optional: Light gray background
    caption={Struct-SQL Prompt Template (QP-CoT) - Two-Shot},
    label={lst:qp_cot_prompt}
]
Answer Repeating the question and evidence, and generating the valid SQLite SQL with a query plan.

###
Two examples of this task are below:
###

### Database Schema:
Table: lists_users
... [Schema details abbreviated] ...

## User Question:
How many more movie lists were created by the user who created the movie list "250 Favourite Films"?

## Hint:
"250 Favourite Films" refers to list_title.

### Response:
**Query Plan**:
** Preparation Steps:**
1. Initialize the process: Start preparing to execute the query.
2. Prepare storage: Set up storage space (registers) to hold temporary results...
3. Open the lists table: Open the lists table so we can read from it.
4. Open the lists_users table: Open the lists_users table so we can read from it.

** Matching and Finding User ID:**
1. Start reading the lists table: Move to the first row in the lists table.
2. Check if the list title matches: Look at the `list_title` column... If it's not "250 Favourite Films," skip this row.
3. Identify the matching row: Store the `user_id` of this list entry. This `user_id` is for the creator of "250 Favourite Films."

** Counting Movie Lists by User:**
1. Prepare to count: If a `user_id` is found from the previous step, prepare to count.
2. Start reading the lists_users table: Move to the first row...
3. Check if the user ID matches: Look at the `user_id` column...
4. Count this match: Increment the count for each row...
5. Move to the next row in lists_users: Go back to the `lists_users` table...
6. Finalize the count: Once all relevant rows have been checked...

** Delivering the Result:**
1. Output the result: Output the final count...
2. End the process: Stop the query execution process.

## SQL Query:
SELECT COUNT(list_id) FROM lists_users WHERE user_id = ( SELECT user_id FROM lists WHERE list_title = '250 Favourite Films' )

## User Question:
What is the director's name of the movie with the highest average rating in 2023, and what is that average rating?

## Hint:
highest average rating refers to MAX(AVG(rating_score)); in 2023 refers to movie_release_year = 2023;

### Response:
**Query Plan**:
** Preparation Steps:**
1. Initialize the process: Start preparing to execute the query.
... [Steps abbreviated] ...
4. Open the ratings table: Open the ratings table so we can read from it.

** Finding the Movie with the Highest Average Rating in 2023 (Subquery Logic):**
1. Start reading the movies and ratings tables.
2. Join movies and ratings tables using their common movie_id column.
3. Filter the joined rows to include only movies where movie_release_year is 2023.
... [Steps abbreviated] ...
6. Select the movie_id of the first (top) result.

** Retrieving Director Name and Average Rating (Main Query Logic):**
1. Start reading the movies and ratings tables again.
... [Steps abbreviated] ...
5. Select the director_name and calculate the AVG(rating_score) for the identified movie.

** Delivering the Result:**
1. Output the director's name and the calculated average rating.
2. End the process: Stop the query execution process.

## SQL Query:
SELECT T1.director_name, AVG(T2.rating_score) FROM movies AS T1 INNER JOIN ratings AS T2 ON T1.movie_id = T2.movie_id WHERE T1.movie_id = (SELECT T1.movie_id FROM movies AS T1 INNER JOIN ratings AS T2 ON T1.movie_id = T2.movie_id WHERE T1.movie_release_year = 2023 GROUP BY T1.movie_id ORDER BY AVG(T2.rating_score) DESC LIMIT 1) GROUP BY T1.director_name


#
# [Target Task Input]
#

### Database Schema:
CREATE TABLE film_text (
    film_id INTEGER not null primary key,
    title TEXT not null,
    description TEXT null
)
... [Target Schema abbreviated] ...

## User Question:
State the name of the category which has the most number of films.

## Hint:
category refers to name; most number of films refers to Max(Count(film_id))

### Response:
**Query Plan**: 
\end{lstlisting}

\section{Unstructured Reasoning Prompt Template (CoT)}
\label{sec:appendix_unstructured_cot}

The \kdcot\ baseline (Listing~\ref{lst:unstructured_cot}) utilizes a 4-shot prompt to elicit free-form natural language reasoning. By requiring the model to 'think step-by-step' without a formal execution plan, this baseline represents the standard unstructured distillation signal.
\begin{lstlisting}[
    basicstyle=\ttfamily\scriptsize,
    breaklines=true,
    frame=single,
    backgroundcolor=\color{gray!5},
    caption={Unstructured CoT Prompt Template (ReasonSQL) - 4-Shot},
    label={lst:unstructured_cot}
]
The goal of the task is to generate a valid SQLite query to answer the question based on schema provided.

###
Few examples of this task are:
###
Schema of the database with sample rows : 

Table: lists_users
Column user_id: column description -> ID related to the user who created the list.
... [Schema details abbreviated] ...

## User Question:
For the list with more than 200 followers, state the title and how long the list has been created?
## Hint: 
more than 200 followers refers to list_followers >200; how long the list has been created refers to SUBTRACT(CURRENT_TIMESTAMP,list_creation_timestamp_utc)
A: Let's think step by step. The SQL query for the given question needs these tables = [lists], so we don't need JOIN.
Plus, it doesn't require nested queries, and we need the answer to the sub-questions = [""].
Also we don't need nested queries. Now to find the lists with more than 200 followers, we need to filter the lists table where list_followers > 200.
Then, we need to select the list_title and calculate the duration since list_creation_timestamp_utc up to the current date.
So the sqlite SQL query will be:
SQL: SELECT list_title , 365 * (strftime('%Y', 'now') - strftime('%Y', list_creation_timestamp_utc)) + 30 * (strftime('%m', 'now') - strftime('%m', list_creation_timestamp_utc)) + strftime('%d', 'now') - strftime('%d', list_creation_timestamp_utc) FROM lists WHERE list_followers > 200

## User Question:
What is the percentage of list created by user who was a subscriber when he created the list?
## Hint:
was a subscriber refers to user_subscriber = 1; percentage refers to DIVIDE(COUNT(user_subscriber = 1),COUNT(list_id))
A: Let's think step by step. The SQL query for the given question needs these tables = [lists_users], so we don't need JOIN.
Also we don't need nested queries. To calculate the percentage, we need to count the total number of lists (COUNT(list_id)) and the number of lists where user_subscriber is 1 (meaning the user was a subscriber).The count of subscribers needs to be cast to REAL to ensure floating-point division for the percentage.
So the sqlite SQL query will be:
SQL: SELECT CAST(SUM(CASE WHEN user_subscriber = 1 THEN 1 ELSE 0 END) AS REAL) * 100 / COUNT(list_id) FROM lists_users

## User Question:
What is the average rating for movie titled 'When Will I Be Loved'?
## Hint:
average rating = DIVIDE((SUM(rating_score where movie_title = 'When Will I Be Loved')), COUNT(rating_score));
A: Let's think step by step. For creating the SQL for the given question, we need to join these tables = [ratings,movies].
First of all, for joining these tables we have to use the common column = [ratings.movie_id = movies.movie_id].
Now, we have to filter the rows where movie_title = 'When Will I Be Loved'.
Then, we have to find the average of the rating_score.
So the sqlite SQL query will be:
SQL: SELECT AVG(T2.rating_score) FROM movies AS T1 INNER JOIN ratings AS T2 ON T1.movie_id = T2.movie_id WHERE T1.movie_title = 'When Will I Be Loved'

## User Question:
For the user who post the list that contained the most number of the movies, is he/she a paying subscriber when creating that list?
## Hint: 
the list that contained the most number of the movies refers to MAX(list_movie_number); user_has_payment_method = 1 means the user was a paying subscriber when he created the list ; \nuser_has_payment_method = 0 means the user was not a paying subscriber when he created the list
A: Let's think step by step. the given question can be solved by knowing the answer to the following sub-questions = [which list has the most number of movies.]
The sqlite SQL query for the sub-question "which list has the most number of movies" is SELECT MAX(list_movie_number) FROM lists
The above query will return the list_movie_number of the list which has the most number of movies.
Now, we have to find the user_has_payment_method of the user who has created the list which has the most number of movies.
To do so, we have to JOIN lists_users and lists table on list_id.
So, the final sqlite SQL query answer to the question the given question is =
SQL: SELECT T1.user_has_payment_method FROM lists_users AS T1 INNER JOIN lists AS T2 ON T1.list_id = T2.list_id WHERE T2.list_movie_number = ( SELECT MAX(list_movie_number) FROM lists )

#
# [Target Task Input]
#

### Database Schema:
CREATE TABLE "country" (
    Name TEXT not null constraint ix_county_Name unique,
    Code TEXT default '' not null primary key,
    ... [Columns abbreviated] ...
)

... [Target Schema abbreviated] ...

## User Question:
Which two countries does the Detroit River flow through? Give the full name of the country.

## Hint:

A: Let's think step by step. 
\end{lstlisting}

\section{ \ourmethod\ Output Examples}
This section provides examples of the \ourmethod\ model's outputs across three levels of SQL complexity: Simple, Moderate, and Challenging. These examples demonstrate the model's ability to generate a human-readable Query Plan that mirrors the logical execution steps of a database engine before producing the final SQLite SQL. Each entry includes the Question ID and Database ID for traceability. The following examples are selected based on the first successful generation encountered for each difficulty tier within the development dataset to ensure an unbiased representation of the model's capabilities.
\begin{tcolorbox}[colback=gray!5, colframe=blue!75!black, title=Struct-SQL Output (Difficulty: Simple)]
\textbf{Question ID:} 5 \\
\textbf{Database ID:} \texttt{california\_schools} \\
\textbf{Question:} How many schools with an average score in Math greater than 400 in the SAT test are exclusively virtual?

\vspace{0.3cm}
\textbf{--- Query Plan ---}
\begin{small}
\begin{verbatim}

** Preparation Steps:**
1. Initialize the process: Start preparing to execute the query.
2. Prepare storage: Set up storage space (registers) to hold temporary results,
   initializing them to NULL.
3. Open the satscores table: Open the satscores table so we can read from it.
4. Open the schools table: Open the schools table so we can read from it.

** Filtering and Joining Tables:**
1. Start reading the satscores table: Move to the first row in the satscores table.
2. Check if the average math score is greater than 400: Look at the `AvgScrMath` column of
   the current row in `satscores`. If it's not greater than 400, skip this row.
3. Identify the matching row: Store the `cds` of this score entry. This `cds` is for schools
   with an average math score greater than 400.
4. Start reading the schools table: Move to the first row in the `schools` table.
5. Check if the school is exclusively virtual: Look at the `Virtual` column of the current row
   in `schools`. If it's not 'F', skip this row.
6. Check if the school matches: Look at the `CDSCode` column of the current row in `schools`. 
   If it matches any `cds` found in the previous step, this school is exclusively virtual
   and has a high average math score.

** Counting Schools:**
1. Prepare to count: If a matching `cds` is found from the previous step, prepare to count.
2. Count this match: Increment the count for each row in `schools` that matches the criteria.
3. Move to the next row in schools: Go back to the `schools` table and move to the next row, 
   repeating the process until all rows are checked.
4. Finalize the count: Once all relevant rows have been checked, finalize the count of schools
   that are exclusively virtual and have an average math score greater than 400.

** Delivering the Result:**
1. Output the result: Output the final count, which is the total number of schools that are
   exclusively virtual and have an average math score greater than 400.
2. End the process: Stop the query execution process.
\end{verbatim}
\end{small}

\textbf{\#\# SQL Query:}
\begin{lstlisting}[language=SQL, basicstyle=\small\ttfamily, breaklines=true]
SELECT COUNT(CDSCode) FROM schools WHERE Virtual = 'F' AND CDSCode IN ( SELECT cds FROM satscores WHERE AvgScrMath > 400 )
\end{lstlisting}
\end{tcolorbox}

\begin{tcolorbox}[colback=gray!5, colframe=orange!75!black, title=Struct-SQL Output (Difficulty: Moderate)]
\textbf{Question ID:} 45 \\
\textbf{Database ID:} \texttt{california\_schools} \\
\textbf{Question:} What is the average writing score of each of the schools managed by Ricci Ulrich? List the schools and the corresponding average writing scores.

\vspace{0.3cm}
\textbf{--- Query Plan ---}
\begin{small}
\begin{verbatim}
** Preparation Steps:**
1. Initialize the process: Start preparing to execute the query.
2. Prepare storage: Set up storage space (registers) to hold temporary results, initializing 
   them to NULL.
3. Open the schools table: Open the schools table so we can read from it.
4. Open the satscores table: Open the satscores table so we can read from it.

** Identifying Schools Managed by Ricci Ulrich:**
1. Start reading the schools table: Move to the first row in the schools table.
2. Check if the administrator's name matches: Look at the `AdmFName1` and `AdmLName1` columns 
   of the current row in `schools`. If they are not "Ricci" and "Ulrich," skip this row.
3. Identify the matching row: Store the `CDSCode` of this school entry. This `CDSCode` is for
   the school managed by Ricci Ulrich.

** Calculating Average Writing Scores:**
1. Prepare to calculate: If a `CDSCode` is found from the previous step, prepare to calculate.
2. Start reading the satscores table: Move to the first row in the `satscores` table.
3. Check if the school code matches: Look at the `cds` column of the current row in `satscores`. 
   If it doesn't match the `CDSCode` found in the "Identifying Schools Managed by Ricci Ulrich"
   step, skip this row.
4. Calculate the average writing score: For each matching row, calculate the average writing
   score using the `AvgScrWrite` column.
5. Move to the next row in satscores: Go back to the `satscores` table and move to the next row,
   repeating the process until all rows are checked.

** Delivering the Result:**
1. Output the result: Output the list of schools managed by Ricci Ulrich and their corresponding
   average writing scores.
2. End the process: Stop the query execution process.
\end{verbatim}
\end{small}

\textbf{\#\# SQL Query:}
\begin{lstlisting}[language=SQL, basicstyle=\small\ttfamily, breaklines=true]
SELECT T1.school, AVG(T2.AvgScrWrite) FROM ( SELECT DISTINCT T1.school, T1.CDSCode FROM schools AS T1 INNER JOIN satscores AS T2 ON T1.CDSCode = T2.cds WHERE T1.AdmFName1 = 'Ricci' AND T1.AdmLName1 = 'Ulrich' ) AS T1 INNER JOIN satscores AS T2 ON T1.CDSCode = T2.cds GROUP BY T1.school
\end{lstlisting}
\end{tcolorbox}

\begin{tcolorbox}[colback=gray!5, colframe=red!75!black, title=Struct-SQL Output (Difficulty: Challenging)]
\textbf{Question ID:} 36 \\
\textbf{Database ID:} \texttt{california\_schools} \\
\textbf{Question:} Under whose administration is the school with the highest number of students scoring 1500 or more on the SAT? Indicate their full names.

\vspace{0.3cm}
\textbf{--- Query Plan ---}
\begin{small}
\begin{verbatim}
## Preparation Steps:
1. Initialize the process: Start preparing to execute the query.
2. Prepare storage: Set up storage space (registers) to hold temporary results, 
   initializing them to NULL.
3. Open the satscores table: Open the satscores table so we can read from it.
4. Open the schools table: Open the schools table so we can read from it.

** Finding the School with the Highest Number of Students Scoring 1500 or More:**
1. Start reading the satscores table: Move to the first row in the satscores table.
2. Check if the score is 1500 or more: Look at the `NumGE1500` column of the current row in 
   `satscores`. If it's not greater than or equal to 1500, skip this row.
3. Identify the matching row: Store the `cds` of this score entry. This `cds` is for the school
   with the highest number of students scoring 1500 or more.

** Retrieving Administrator Full Names:**
1. Prepare to retrieve administrator names: If a `cds` is found from the previous step, prepare
   to retrieve administrator names.
2. Start reading the schools table: Move to the first row in the `schools` table.
3. Check if the school code matches: Look at the `CDSCode` column of the current row in `schools`.
   If it doesn't match the `cds` found in the "Finding the School with the Highest Number of 
   Students Scoring 1500 or More" step, skip this row.
4. Retrieve administrator names: For each matching row, retrieve the `AdmFName1`, `AdmLName1`,
   `AdmFName2`, `AdmLName2`, `AdmFName3`, and `AdmLName3` columns, which represent the full 
   names of the administrators.
5. Move to the next row in schools: Go back to the `schools` table and move to the next row,
   repeating the process until all rows are checked.

** Delivering the Result:**
1. Output the administrator full names: Output the full names of the administrators for the 
   school with the highest number of students scoring 1500 or more.
2. End the process: Stop the query execution process.
\end{verbatim}
\end{small}

\textbf{\#\# SQL Query:}
\begin{lstlisting}[language=SQL, basicstyle=\small\ttfamily, breaklines=true]
SELECT T1.AdmFName1, T1.AdmLName1, T1.AdmFName2, T1.AdmLName2, T1.AdmFName3, T1.AdmLName3 FROM schools AS T1 INNER JOIN satscores AS T2 ON T1.CDSCode = T2.cds WHERE T2.NumGE1500 = ( SELECT MAX(NumGE1500) FROM satscores )
\end{lstlisting}
\end{tcolorbox}

\end{document}